\documentclass{article}
\usepackage[utf8]{inputenc}
\usepackage{main}
\usepackage{microtype}
\usepackage{subcaption}
\usepackage{graphicx}
\usepackage{times}
\usepackage{latexsym}
\usepackage{amsmath}
\usepackage{float}
\usepackage{footnote}
\usepackage{enumitem}
\usepackage{bm}
\usepackage{arydshln}
\usepackage{booktabs}
\usepackage{multicol}
\usepackage{multirow}
\usepackage{color}
\usepackage{xcolor}     
\usepackage{colortbl}
\usepackage{bbding}
\usepackage{makecell}
\usepackage{mathtools}
\usepackage{imakeidx}
\usepackage{longtable}
\usepackage{wrapfig}
\makeindex
\usepackage{arydshln}
\usepackage{lipsum}
\usepackage{natbib}
\usepackage[toc]{multitoc}
\usepackage[edges]{forest}
\usepackage[normalem]{ulem}
\definecolor{mydarkblue}{rgb}{0,0.08,0.45}
\usepackage[colorlinks=true,linkcolor=mydarkblue,citecolor=mydarkblue,filecolor=mydarkblue,urlcolor=mydarkblue]{hyperref}
\usepackage{cleveref}
\usepackage{CJKutf8}
\usepackage{awesomebox} 
\usepackage{bbding}
\usepackage[most]{tcolorbox}
\usepackage{booktabs}
\usepackage{geometry}
\definecolor{wkblue}{rgb}{0.2, 0.3, 0.6}
\definecolor{meta-color}{rgb}{0.5, 0.5, 0.5}
\usepackage{amsmath}
\usepackage{enumitem}

\usepackage[tikz]{bclogo}
\usepackage[framemethod=tikz]{mdframed}
\definecolor{bgblue}{RGB}{245,243,253}
\definecolor{ttblue}{RGB}{91,194,224}

\mdfdefinestyle{mystyle}{%
  rightline=true,
  innerleftmargin=10,
  innerrightmargin=10,
  outerlinewidth=3pt,
  topline=false,
  rightline=true,
  bottomline=false,
  skipabove=\topsep,
  skipbelow=\topsep
}

\newtcolorbox{myboxi}[1][]{
  breakable,
  title=#1,
  colback=red!5,
  colbacktitle=red!5,
  coltitle=black,
  fonttitle=\bfseries,
  bottomrule=0pt,
  toprule=0pt,
  leftrule=2pt,
  rightrule=2pt,
  titlerule=0pt,
  arc=0pt,
  outer arc=0pt,
  colframe=red,
}

\newtcolorbox{myboxnote}[1][]{
  breakable,
  title=#1,
  colback=orange!0,
  colbacktitle=orange!0,
  coltitle=black,
  fonttitle=\bfseries,
  bottomrule=0pt,
  toprule=0pt,
  leftrule=2pt,
  rightrule=2pt,
  titlerule=0pt,
  arc=0pt,
  outer arc=0pt,
  colframe=orange,
}

\newtcolorbox{myboxii}[1][]{
  breakable,
  freelance,
  title=#1,
  colback=white,
  colbacktitle=white,
  coltitle=black,
  fonttitle=\bfseries,
  bottomrule=0pt,
  boxrule=0pt,
  colframe=white,
  overlay unbroken and first={
  \draw[red!75!black,line width=3pt]
    ([xshift=5pt]frame.north west) -- 
    (frame.north west) -- 
    (frame.south west);
  \draw[red!75!black,line width=3pt]
    ([xshift=-5pt]frame.north east) -- 
    (frame.north east) -- 
    (frame.south east);
  },
  overlay unbroken app={
  \draw[red!75!black,line width=3pt,line cap=rect]
    (frame.south west) -- 
    ([xshift=5pt]frame.south west);
  \draw[red!75!black,line width=3pt,line cap=rect]
    (frame.south east) -- 
    ([xshift=-5pt]frame.south east);
  },
  overlay middle and last={
  \draw[red!75!black,line width=3pt]
    (frame.north west) -- 
    (frame.south west);
  \draw[red!75!black,line width=3pt]
    (frame.north east) -- 
    (frame.south east);
  },
  overlay last app={
  \draw[red!75!black,line width=3pt,line cap=rect]
    (frame.south west) --
    ([xshift=5pt]frame.south west);
  \draw[red!75!black,line width=3pt,line cap=rect]
    (frame.south east) --
    ([xshift=-5pt]frame.south east);
  },
}

\usepackage{fancyhdr} 
\usepackage{blindtext} 

\DeclareCaptionFont{black}{\color{black}}

\definecolor{myblue}{rgb}{0.9, 0.1, 0.94}
\definecolor{mygreen}{rgb}{0.64, 0.56, 0.88}
\definecolor{myyellow}{rgb}{0.68, 0.6, 0.1}
\definecolor{fancygreen}{rgb}{0.33, 0.68, 0.20}
\definecolor{salmon}{rgb}{0.94, 0.52, 0.49}
\definecolor{tablegreen}{rgb}{0.82, 0.94, 0.75}
\definecolor{tableblue}{rgb}{0.81, 0.90, 0.94}
\definecolor{tablered}{rgb}{0.97, 0.85, 0.85}
\definecolor{tableorange}{rgb}{0.96, 0.85, 0.81}
\definecolor{TiffanyBlue}{rgb}{0.0, 0.68, 0.67}

\newenvironment{itemize*}%
 {\leftmargini=10pt\begin{itemize}%
  \setlength{\itemsep}{0pt}%
  \setlength{\parskip}{0pt}%
  }%
 {\end{itemize}}
\newenvironment{enumerate*}%
 {\begin{enumerate}%
  \setlength{\itemsep}{0pt}%
  \setlength{\parskip}{0pt}}%
 {\end{enumerate}}

\usepackage{xcolor}
\usepackage{listings}

\newcommand\JSONnumbervaluestyle{\color{blue}}
\newcommand\JSONstringvaluestyle{\color{red}}

\newif\ifcolonfoundonthisline

\makeatletter

\lstdefinestyle{json}
{
  showstringspaces    = false,
  keywords            = {false,true},
  alsoletter          = 0123456789.,
  morestring          = [s]{"}{"},
  stringstyle         = \ifcolonfoundonthisline\JSONstringvaluestyle\fi,
  MoreSelectCharTable =%
    \lst@DefSaveDef{`:}\colon@json{\processColon@json},
  basicstyle          = \ttfamily,
  keywordstyle        = \ttfamily\bfseries,
}

\newcommand\processColon@json{%
  \colon@json%
  \ifnum\lst@mode=\lst@Pmode%
    \global\colonfoundonthislinetrue%
  \fi
}

\lst@AddToHook{Output}{%
  \ifcolonfoundonthisline%
    \ifnum\lst@mode=\lst@Pmode%
      \def\lst@thestyle{\JSONnumbervaluestyle}%
    \fi
  \fi
  \lsthk@DetectKeywords%
}

\lst@AddToHook{EOL}%
  {\global\colonfoundonthislinefalse}

\makeatother

\usepackage{etoolbox}
\usepackage{natbib}
\usepackage{url}
\newcounter{bibcount}
\makeatletter
\patchcmd{\@lbibitem}{\item[}{\item[\hfil\stepcounter{bibcount}{[\thebibcount]}}{}{}
\renewcommand\NAT@bibsetup%
  [1]{\setlength{\leftmargin}{\bibhang}\setlength{\itemindent}{-\parindent}%
      \setlength{\itemsep}{\bibsep}\setlength{\parsep}{\z@}}
\makeatother

\newcommand*\samethanks[1][\value{footnote}]{\footnotemark[#1]}

\author{%
Zhongzhen Huang$^{1,3}$\thanks{~~Co-first authors}\space\space\space\space Gui Geng$^{3}$\samethanks\space\space\space\space Shengyi Hua$^{1,3}$\samethanks\space\space\space\space Zhen Huang$^{4}$\samethanks\space\space\space\space Haoyang Zou$^{4}$\samethanks\\
\textbf{Shaoting Zhang$^{1}$\thanks{~~Corresponding author}\space\space\space Pengfei Liu$^{1, 2, 4}$\samethanks\space\space\space Xiaofan Zhang$^{1,3}$\samethanks}\\
$^1$Shanghai Jiao Tong University , $^2$SII, \\ $^3$SPIRAL Lab, $^4$Generative AI Research Lab (GAIR),}

\begin{document}

\title{O1 Replication Journey – Part 3:\\
Inference-time Scaling for Medical Reasoning}

\maketitle
\thispagestyle{fancy}
\fancyhead{}
\lhead{}
\lhead{\includegraphics[height=0.67cm]{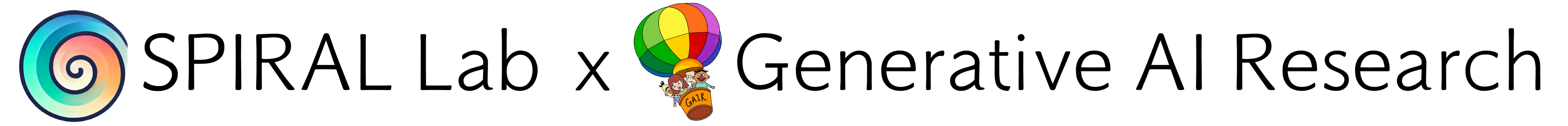}}
\renewcommand{\headrulewidth}{0pt}
\setlength{\headsep}{0mm}

\begin{abstract}

Building upon our previous investigations of O1 replication (Part 1: Journey Learning~\citep{qin2024o1replicationjourneystrategic} and Part 2: Distillation~\citep{huang2024o1}), this work explores the potential of inference-time scaling in large language models (LLMs) for medical reasoning tasks, ranging from diagnostic decision-making to treatment planning. Through extensive experiments on medical benchmarks of varying complexity (MedQA, Medbullets, and JAMA Clinical Challenges),
our investigation reveals several key insights: (1) Increasing inference time does lead to improved performance. With a modest training set of 500 samples, our model yields substantial performance improvements of 6\%-11\%.
(2) Task complexity directly correlates with the required length of reasoning chains, confirming the necessity of extended thought processes for challenging problems;
(3) The differential diagnoses generated by our model adhere to the principles of the hypothetico-deductive method, producing a list of potential conditions that may explain a patient's symptoms and systematically narrowing these possibilities by evaluating the evidence.
These findings demonstrate the promising synergy between inference-time scaling and journey learning in advancing LLMs' real-world clinical reasoning capabilities.
Resources are available at \url{https://github.com/SPIRAL-MED/Ophiuchus}, which is a part of O1 Journey Project.

\end{abstract}
\vspace{-10pt}
\begin{figure}[h]
\centering
\scalebox{1}{
\includegraphics[width=\linewidth]{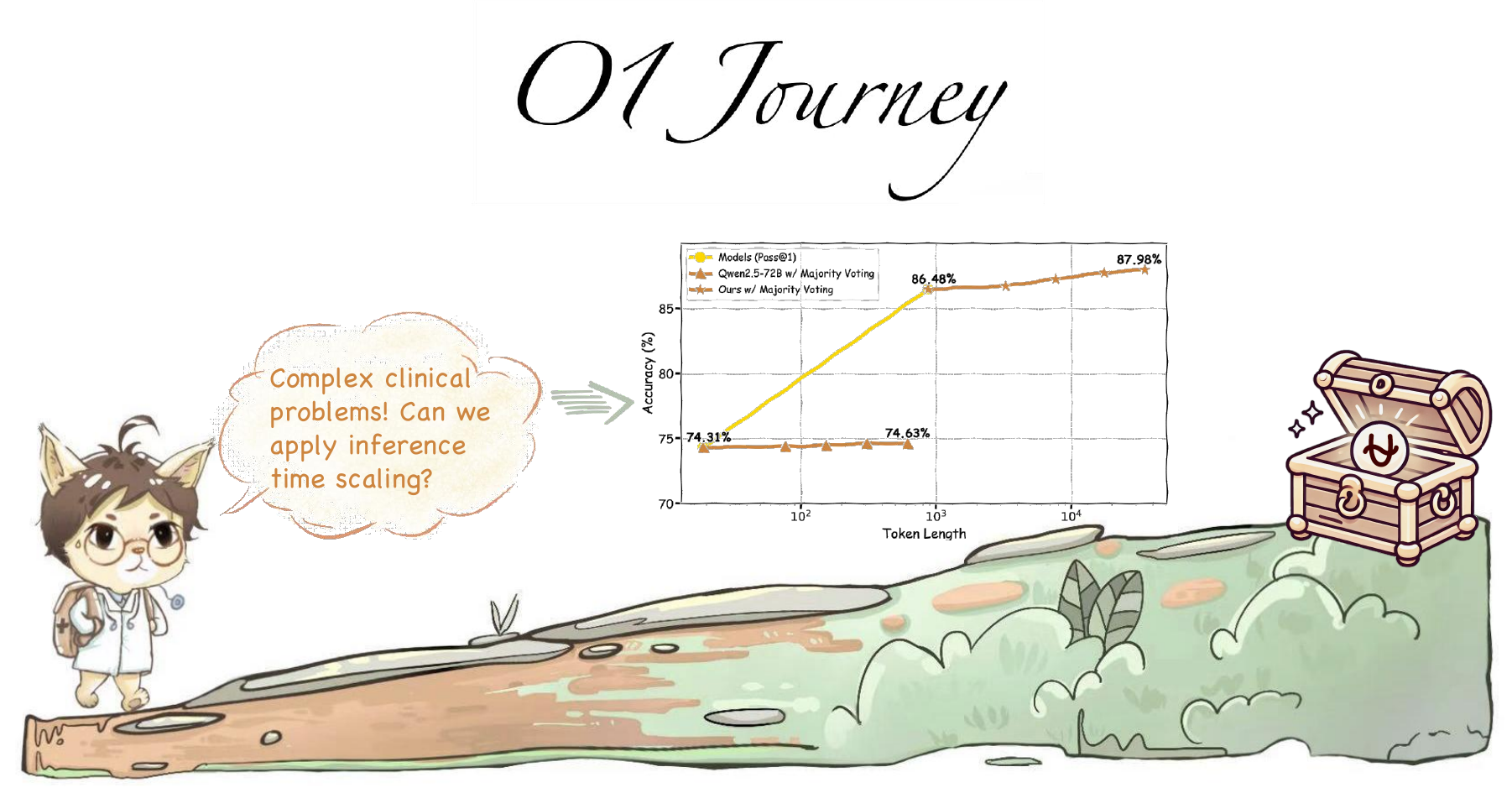}
}
\caption{Illustration of our O1 replication journey in the medical field. \includegraphics[scale=0.1]{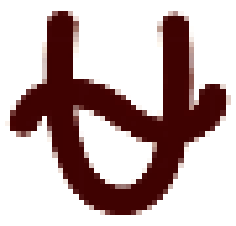} which aims to develop systems capable of deep scientific thinking, ultimately enabling AI-driven breakthroughs in medical domains.}
\label{fig:intro}
\end{figure}

\newpage
\pagestyle{fancy}
\lhead{}
\renewcommand{\headrulewidth}{0.7pt}
\setlength{\headsep}{5mm}


\clearpage

\section{Introduction}

Medicine is an endeavor that fundamentally involves complex reasoning, spanning tasks from diagnostic decision-making to treatment planning. Intricate reasoning is particularly crucial in medical scenarios where patient outcomes depend on understanding multifactorial conditions~\cite{singhal2023large, qiu2024towards}. The process of differential diagnosis~\cite{seller2011differential} exemplifies this complexity, requiring physicians to generate a list of potential diagnoses and methodically narrow it down by evaluating clinical findings and excluding options that do not align with the evidence. This process demands not only deep clinical knowledge but also the ability to draw logical inferences and evaluate multiple hypotheses based on available evidence.
In recent years, the emergence of large language models (LLMs) holds significant promise for advancing clinical applications. However, the complexity of medical reasoning poses unique challenges that traditional scaling methods—such as increasing model parameters or training data volume~\cite{kaplan2020scaling, brown2020language, chowdhery2023palm}—struggle to address effectively. Recent studies~\cite{snell2024scaling, openai2024reasoning} have demonstrated that scaling inference time can lead to more efficient improvements in LLM performance. This approach, known as \textit{inference-time scaling}, allows more processing time for complex tasks, enabling step-by-step problem-solving and iterative refinement of the reasoning process. OpenAI's O1~\citep{openai2024reasoning} particularly emphasizes this strategy by scaling inference time to generate long thoughts for complex reasoning.

While previous studies~\cite{huang2024o1, jiang2024technical, zhang2024llamaberry, zhang2024o1coder, k0math, skyworko1, qwq-32b-preview} have validated inference-time scaling in various domains such as mathematical reasoning, its application to medical scenarios presents unique opportunities and challenges. Building on our previous work on journey learning~\cite{qin2024o1replicationjourneystrategic}, we posit that complex clinical scenarios—requiring integration of established knowledge with nuanced patient histories and comorbidities—necessitate even longer reasoning processes. Our initial explorations demonstrated significant improvements (Figure~\ref{fig:intro}) when scaling inference time to account for multiple clinical factors and iteratively refine diagnostic strategies.
This work examines how LLMs with \textit{inference-time scaling} adapt to varying task complexities in medicine, focusing on two key aspects: (1) identifying optimal scenarios for assessing performance variations with \textit{inference-time computing}, and (2) developing effective methods to synthesize supervised fine-tuning data for generating extended reasoning chains. To validate our approach, we utilize public medical benchmarks proposed by~\cite{chen2024benchmarking}, encompassing challenging clinical cases and medical licensing examinations. A key challenge lies in enabling effective journey learning during inference. Building on previous work~\cite{huang2024o1}, we employ a straightforward yet effective knowledge distillation approach~\cite{hinton2015distilling} from GPT-series models.
Key findings from our experiments include the following: (1) majority voting provides a straightforward approach to augment inference-time computation, though its effectiveness is constrained in complex scenarios;
(2) effective scaling of inference time, however, depends on the sufficient capacity of LLMs, otherwise, such efforts are likely to be in vain; (3) tasks with greater complexity necessitate longer reasoning processes, reinforcing the need for extended thought chains as task difficulty increases; and (4) removing multiple-choice options and encouraging free-form responses unlocks the potential for medical journey learning, fostering nuanced clinical reasoning.

Looking ahead, the integration of \textit{inference-time scaling} into clinical applications presents both opportunities and challenges. Our work underscores the potential of this approach to address complex medical tasks, but it also highlights the need for continued exploration and innovation. By releasing key findings, distilled datasets, and experimental methodologies, we aim to contribute to the broader AI research community and foster collaborative advancements. Furthermore, we advocate for a research ethos rooted in transparency, originality, and rigorous evaluation, particularly as AI systems become more deeply embedded in critical domains like healthcare. While the distillation strategy and its implications warrant further scrutiny, we hope this study catalyzes future investigations into \textit{inference-time scaling} and its capacity to enhance the reasoning capabilities of LLMs. Ultimately, our goal is to inspire new methodologies and applications that bridge the gap between computational innovation and practical medical impact, ensuring better outcomes for patients and practitioners alike.


%



\section{Exploration Process}
\label{sec:exp_process_begin}
The conclusion derived from analyzing examples provided by OpenAI~\footnote{\href{https://openai.com/index/learning-to-reason-with-llms}{https://openai.com/index/learning-to-reason-with-llms}} demonstrates that \textbf{as the difficulty increases, the inference time tends to grow proportionally}~\cite{qin2024o1replicationjourneystrategic}. This suggests that higher-difficulty problems require more reasoning steps, which in turn necessitate a longer inference time.  Although the exact mechanisms through which \textit{inference-time scaling} enhances problem-solving remain underexplored, it is evident that \textit{inference-time scaling} contributes significantly to identifying and analyzing key information. This phenomenon is particularly critical in medical domains, where clinicians require much time to process data from multiple sources and modalities when diagnosing conditions, making prognostic evaluations, and determining treatment plans. In the following parts, we document our exploration of evaluating the utility of \textit{inference-time scaling} in addressing complex, domain-specific challenges in medicine.

\subsection{Benchmark Overview}
To demonstrate the effectiveness of \textit{inference-time scaling} in addressing medical problems, we selected three benchmarks in~\cite{chen2024benchmarking} for our experiments: the JAMA Clinical Challenge (JAMA), Medbullets, and MedQA. These benchmarks encompass challenging real-world clinical cases from various medical domains as well as medical licensing exams of different difficulty levels. 
The JAMA dataset includes 1,524 examples collected from the JAMA Network Clinical Challenge~\footnote{\href{https://jamanetwork.com/collections/44038/clinical-challenge}{https://jamanetwork.com/collections/44038/clinical-challenge}} archive, spanning the past decade (July 2013–October 2023) and covering 13 medical domains. The examples are based on complicated clinical scenarios involving patient history, family history, laboratory results, physical/radiology/cardiology analysis, etc., and hence require more sophisticated understanding and reasoning to ``arrive at a correct diagnosis''. At this stage, our primary goal is to examine the effectiveness of \textit{inference-time scaling} on complex tasks. Therefore, we focused on cases that o1-mini struggles with. Specifically, we utilized a streamlined subset of JAMA containing 646 cases, half of which are challenging for o1-mini, for evaluation.

The Medbullets and MedQA datasets in use are based on the National Medical Board Examination in the United States or the United States Medical Licensing Examination (USMLE). Medbullets is an online platform that provides medical study resources. The dataset focuses on Medbullets \textit{Step 2/3}~\footnote{\href{https://step2.medbullets.com/}{https://step2.medbullets.com/}} which serves USMLE \textit{Step 2\&3}~\footnote{\href{https://www.usmle.org/step-exams/step-2-ck}{https://www.usmle.org/step-exams/step-2-ck}} type questions. The resolution of the questions in \textit{Step 2/3} demands the application of medical knowledge and clinical reasoning rather than relying solely on textbook knowledge (\textit{Step 1} questions). The dataset consists of 308 examples posted on the X (formerly Twitter)~\footnote{\href{https://x.com/medbullets}{https://x.com/medbullets}} platform between April 2022 and December 2023. Each comprises a case description, a question, five answer choices, and an explanation that explains each option. MedQA also includes questions from the Medbullets website but without the aforementioned explanations. The test set includes 679 \textit{Step 1}~\footnote{\href{https://step1.medbullets.com/}{https://step1.medbullets.com/}} questions and 594 \textit{Step 2/3} questions, obtained in March 2021, ensuring that there is no overlap with the Medbullets dataset. The involvement of these two datasets provides insights into whether \textit{inference-time scaling} helps or hinders tackling medical tasks of various difficulty levels, and how it takes effect.

\subsection{Journey Learning Data Synthesis}

To enable LLMs to perform journey learning during the problem-solving process, we need to construct a collection of high-quality demonstration data that exhibits this behavior. Building on prior efforts, we adopted a distillation-based approach for producing high-quality data. In the era of LLMs, instructing weaker models using stronger ones is common practice for advancing models. Recent research~\cite {taori2023alpaca, xu2023wizardlm, gunasekar2023textbooks} highlights that fine-tuning with high-quality data synthesized from proprietary models can achieve remarkable outcomes. In our exploration, we utilized o1 to collect journey learning data, synthesizing two types of long-form data: \texttt{LongStep} and \texttt{LongMonolog}.

\textbf{LongStep}: Upon analyzing responses from o1 and GPT-4o, we observed that o1 generates longer solution steps that include more detailed processes for analyzing key information. Given o1’s exceptional performance and sophisticated reasoning capabilities, we extracted its solution steps to train LLMs to emulate this behavior, producing more thorough and detailed solutions.

\textbf{LongMonolog}: In practice, the examples provided by OpenAI often exhibit a relatively flexible, sometimes colloquial, ``inner monolog'' style. However, the internal thought processes of o1 cannot be directly accessed because it summarizes these processes before presenting them to the user. Despite this, its summarized internal thoughts are still valuable, as they outline the key stages of problem-solving. That said, the summarized thoughts are not directly suitable for training. Inspired by~\cite{huang2024o1}, we designed prompts to instruct o1-preview to expand its summarized thoughts into a long-form reasoning. Careful prompt design is essential, as o1 imposes restrictions on accessing its internal reasoning processes. The resulting output adheres to specific guidelines, ensuring that the solutions resemble an inner monologue, are highly detailed and reflective, including self-corrections, and exhibit extended reasoning. After collecting labeled data, we performed further preprocessing to ensure data quality and standardization of output formats. 

At the current stage, our primary goal is to evaluate the role of \textit{inference-time scaling} in addressing medical problems. Our objective is not to directly perform differential diagnosis, which we acknowledge as extremely difficult given the limited information and resources available. In real-world scenarios, differential diagnosis aligns with the principles of the hypothetico-deductive method, where potential diseases or conditions are treated as hypotheses that clinicians evaluate to determine their validity. To simplify the task, we adopt multiple-choice datasets in this section, allowing potential diagnoses (the ``differential'') can be predefined to guide the model in generating hypotheses. We opt not to use in-house data, as real-world clinical scenarios often contain a substantial amount of irrelevant information that can interfere with reasoning. This poses a great challenge for current models. In contrast, public benchmarks simplify the problem and eliminate some of this interference. Moreover, the choices from MedQA and JAMA are carefully designed to be similar and plausible options. In analyzing these options to determine the final answer, the process closely mirrors the thought process involved in clinical diagnosis. Moreover, the key factor in data selection was the length of the problem-solving process. We excluded cases with a short thought process. Collectively, we assembled a training dataset consisting of 500 examples, with 350 drawn from the training set of MedQA~\cite{jin2020disease} and 150 from the remaining set of the JAMA Clinical Challenge. The average length is 729 for the curate \texttt{LongStep} dataset and 1,223 for the \texttt{LongMonolog} dataset, respectively. Examples of our distilled data are presented in \Cref{fig:case_problem,fig:case_step,fig:case_monolog}.

\begin{table*}[htb]
    \centering
    \renewcommand\arraystretch{1.1}
    \resizebox{\linewidth}{!}{
    \begin{tabular}{@{}l c c |cc|cc|cc}
    \toprule
     \multirow{2}{*}{Model Name} & \multirow{2}{*}{\makecell[c]{Param.\\Size}} &\multirow{2}{*}{\makecell[c]{ Mean Acc.}}  & \multicolumn{2}{c|}{JAMA (646)}  & \multicolumn{2}{c|}{Medbullets (308)} & \multicolumn{2}{c}{MedQA (1273)} \\
     \cmidrule{4-9}
     & &  & Acc. & \#Avg. Token & Acc. & \#Avg. Token & Acc. & \#Avg. Token \\
     \midrule
     \multicolumn{9}{c}{\textit{\textbf{Proprietary}}} \\
     \midrule
     GPT-4o & - & 80.01 & 63.77 & - &77.92 & - &88.76 & -\\
     GPT-4o-Vanilla CoT & - & 81.83 & 63.77 & 335 & 81.68 & 323 &91.04 & 300\\
     o1-mini & - & 77.45 & 50.00 & - & 80.51 & - &90.65 & -\\
     o1-preview & - & 87.95 & 73.21 & - & 89.28 & - & 95.12 & -\\
     \midrule
    \multicolumn{9}{c}{\textit{\textbf{Open Source Models}}} \\
     \midrule
     \rowcolor{yellow!8}Qwen2.5 & 7B & 51.23 & 42.41 & - &46.75 & - &56.79 & -\\
     \rowcolor{yellow!8}Qwen2.5-Vanilla CoT & 7B & 49.61 & 40.40 & 371 &46.42 & 353 &55.06 & 344\\
     \rowcolor{yellow!8}InternLM2.5 & 7B & 45.84 & 39.78 & - &43.18 & - &49.56 & - \\
    \rowcolor{yellow!8}InternLM2.5-Vanilla CoT & 7B & 42.47 & 34.05 & 303 &38.63 & 273 &48.15 & 291 \\
    \rowcolor{yellow!8}LLama3.1 & 8B & 55.71 & 46.67 & - &49.43 & - &61.82 & - \\
    \rowcolor{yellow!8}InternLM2.5 &20B & 51.90 & 43.34 & - &49.02 & - &56.95 & - \\
    \rowcolor{yellow!8}InternLM2.5-Vanilla CoT &20B & 51.18 & 42.41 & 312 &46.01 & 330 &56.87 & 323 \\

    \rowcolor{red!8}Qwen2.5 &32B & 64.20 & 49.84 & - &59.74 & - &72.58 & - \\
    \rowcolor{red!8}Qwen2.5-Vanilla CoT &32B & 65.86 & 50.92 & 351 &61.68 & 332 &74.46 & 329 \\
    \rowcolor{red!8}LLama3.1 &70B & 71.39 & 59.59 & - &67.85 & - &78.24 & - \\
    \rowcolor{red!8}LLama3.1-Vanilla CoT &70B & 73.59 & 57.27 & 529 &66.55 & 496 &83.11 & 477 \\
    \rowcolor{red!8}Qwen2.5 &72B & 65.82 & 50.15 & - &63.63 & - &74.31 & - \\
    \rowcolor{red!8}Qwen2.5-Vanilla CoT &72B & 69.10 & 50.15 & 435 &65.58 & 387 &79.57 & 375 \\
     \midrule
    \multicolumn{9}{c}{\textit{\textbf{Supervised Fine-tuning with Vanilla CoT}}} \\
    \midrule
    \rowcolor{orange!8}Qwen2.5-\texttt{CoT SFT} &32B & 67.13 & 52.16 & 383 & 61.03 & 340 & 76.19 & 323 \\
    \rowcolor{orange!8}LLama3.1-\texttt{CoT SFT} &70B & 74.48 & 56.03 & 386 & 72.07 & 342 & 84.44 & 333 \\
    \rowcolor{orange!8}Qwen2.5-\texttt{CoT SFT} &72B & 70.94 & 53.09 & 402 & 67.20 & 355 & 80.91 & 346 \\
     \midrule
     \multicolumn{9}{c}{\textit{\textbf{Ours (Journey Learning)}}} \\
     \midrule
     \rowcolor{cyan!8}Qwen2.5-\texttt{LongStep} &32B & 70.08 & 56.34 & 759 & 66.23 & 645 & 78.00 & 615 \\
     \rowcolor{cyan!8}Qwen2.5-\texttt{LongMonolog} &32B & 70.23 & 53.71 & 1098 & 68.50 & 1023 & 79.02 & 997 \\
    \rowcolor{cyan!8}LLama3.1-\texttt{LongStep} &70B & 76.59 &\textbf{60.21} & 819 & 74.67 & 721 & 85.38 & 669 \\
    \rowcolor{cyan!8}LLama3.1-\texttt{LongMonolog} &70B & \textbf{77.36} & \underline{59.44} & 1153 & \textbf{77.27} & 1029 & \textbf{86.48} & 953 \\
    \rowcolor{cyan!8}Qwen2.5-\texttt{LongStep} &72B & 75.51 & 58.66 & 762 & 72.07 & 692 & 84.91 & 631 \\
    \rowcolor{cyan!8}Qwen2.5-\texttt{LongMonolog} &72B & \underline{77.18} & 59.28 & 1076 &\underline{76.29}  & 917 &\textbf{86.48} & 873 \\ 
    \bottomrule     
    \end{tabular}
    }
    \caption{Performance comparison of proprietary, open-source, and supervised fine-tuned models across three medical benchmarks. The metrics include accuracy (Acc.) and average output token length (Avg. Token). Mean Acc. represents the weighted average across the three datasets. The best score excluding closed-source APIs was bolded, and the second-best score was underlined.}
    \label{tab:main}
\end{table*}

\section{Experiments}

\subsection{Implementation Details}
In our pilot studies, we found that current open-source LLMs still lag significantly behind commercial closed-source APIs. Considering resource constraints and the need to share our results with the community promptly, we selected Qwen2.5-32B-Instruct~\cite{qwen2.5}, Qwen2.5-72B-Instruct~\cite{qwen2.5} and LLama3.1-70B-Instruct~\cite{llama3} as our base models due to their foundational capabilities in medicine. This foundational knowledge ensures a solid starting point for subsequent long reasoning processes and improvements. We utilized Llama-Factory~\cite{zheng2024llamafactory} to perform instruction tuning on the LLMs with LoRA~\cite{hu2021lora}. Additionally, we employed DeepSpeed optimization~\cite{rasley2020deepspeed} with ZeRO-3 configuration. Following the methodology outlined in~\cite{qin2024o1replicationjourneystrategic, huang2024o1}, we set the number of training epochs to three. All LLMs were fine-tuned on {8} NVIDIA A800 GPUs using a learning rate of $1 \times 10^{-4}$ and a batch size of {8}.

\subsection{Main Results}
In this study, we present a comprehensive performance comparison of various methods on the evaluation benchmarks listed in Table~\ref{tab:main}. The results include performance metrics for proprietary APIs, open-source baselines, and several models fine-tuned on our synthesized data. In Table~\ref{tab:main}, the rows with the ``-Vanilla CoT'' suffix indicate models or APIs that solve problems step by step using Chain-of-Thought prompts according to~\cite{wei2022chain}. We selected models that benefit from introducing a reasoning process (highlighted in red) for further experiments. The rows marked in orange (``-\texttt{CoT SFT}'' suffix) show the result of models trained with vanilla CoT from GPT-4o.  The rows marked in blue correspond to models, with the ``-\texttt{LongStep}'' suffix and ``-\texttt{LongMonolog}'' suffix, fine-tuned on our two journey learning datasets. Additionally, we measure the average token count of each model's outputs via ``tiktoken''.~\footnote{\href{https://github.com/openai/tiktoken}{https://github.com/openai/tiktoken}} This is crucial for evaluating the effectiveness of inference time scaling schemes. The results demonstrate that these methods significantly improve model performance even with minimal training data. These findings highlight the effectiveness of inference-time computing in enhancing the complex reasoning capabilities of LLMs for addressing medical challenges. A more detailed analysis can be seen in the next sections.

\begin{figure}[tb]
\centering
\scalebox{1}{
\includegraphics[width=\linewidth]{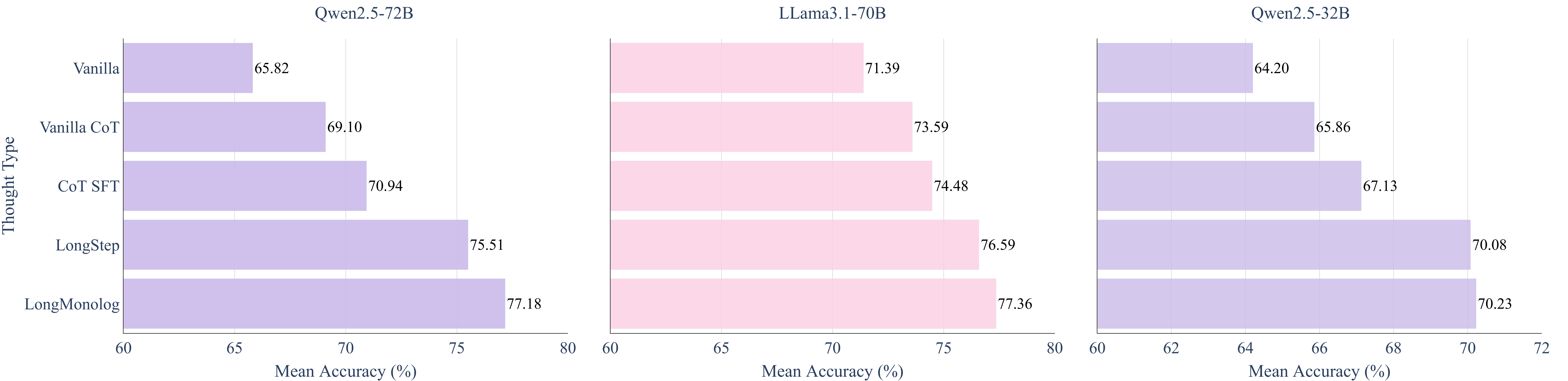}
}
\caption{Weighted mean accuracy of Qwen2.5-72B-Instruct, LLama3.1-70B, and Qwen2.5-32B across three datasets using distinct strategies.}
\label{fig:mean_acc}
\end{figure}

\subsection{Does Inference-time Scaling Help?}
To intuitively illustrate the contribution of \textit{inference-time computing}, we present the accuracy of Qwen2.5-72B, LLama3.1-70B, and Qwen2.5-32B across three datasets using distinct strategies: vanilla usage, vanilla CoT prompting, CoT SFT, long step SFT, and long monolog SFT. As shown in Figure~\ref{fig:mean_acc}, each strategy significantly improves overall accuracy. Notably, Qwen2.5-72B achieves gains of +3.28\%, +5.12\%, +9.69\%, and +11.36\% for the respective strategies, strongly supporting the hypothesis that incorporating structured thought processes in inference time enhances the ability of powerful models to address complex medical problems. 

A key observation is that \textbf{more inference time leads to enhanced performance.} For instance, when Qwen2.5-72B employs step-by-step reasoning, whether via vanilla CoT or CoT fine-tuning, the output token length ranges from 300 to 500 tokens, resulting in about 5\% increase in mean accuracy. In contrast, under the journey learning settings, which include long step and long monolog fine-tuning, the token count extends to approximately 1,000, yielding improvements of about 10\%. A similar trend is evident for Qwen2.5-32B and LLama3.1-70B, as depicted in the remaining figures of Figure~\ref{fig:mean_acc}. Specifically, LLama3.1-70B achieves improvements of +2.20\%, +3.09\%, +5.20\%, and +5.97\%, while Qwen2.5-32B shows gains of +1.66\%, +2.93\%, +5.88\%, and +6.03\% across the three reasoning strategies.

When comparing \texttt{LongStep} and \texttt{LongMonolog}, it remains challenging to determine which consistently delivers superior performance. Based on current experimental data, long monolog demonstrates higher accuracy on Medbullets and MedQA but does not maintain its advantage in JAMA. For instance, Qwen2.5-32B achieves 56.34\% accuracy in JAMA with \texttt{LongStep} but only 53.71\% with \texttt{LongMonolog}. The limitation in Qwen2.5-32B's self-reflective reasoning may stem from its inability to construct a complete thought leading to correctness. As shown in \Cref{fig:case_problem_32,fig:case_step_32,fig:case_monolog_32}, extended steps result in correct answers, whereas redundant reflections sometimes lead to errors (red texts in Figure~\ref{fig:case_monolog_32}). This finding suggests an assumption that long thought processes during inference time can aid in answering complex medical questions but require sufficient domain knowledge.

\begin{figure}[t]
\centering
\scalebox{1}{
\includegraphics[width=\linewidth]{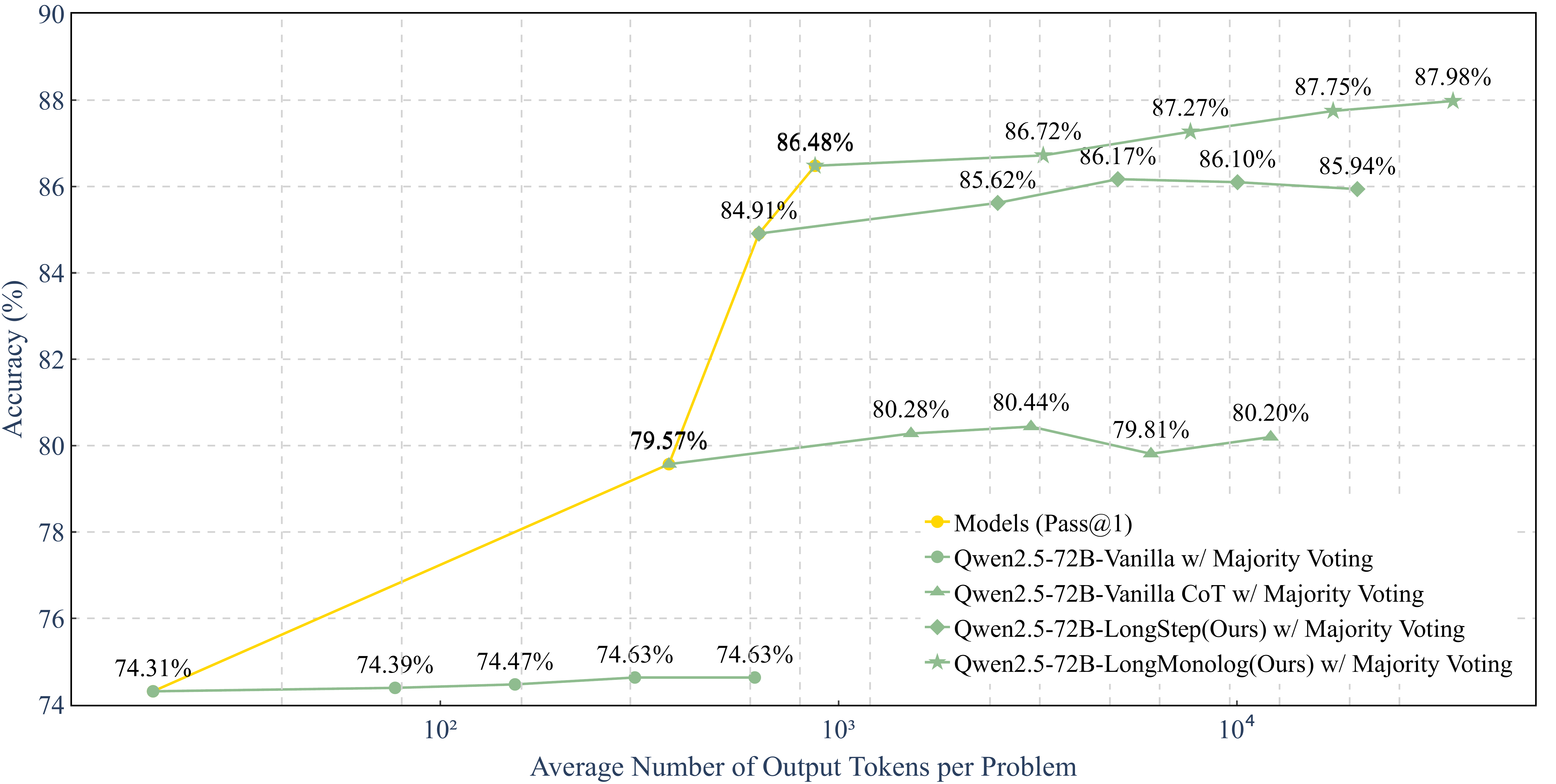}
}
\caption{The Accuracy of Qwen2.5-72B-Series on MedQA with inference-time scaling.}
\label{fig:mv_line_2}
\end{figure}

\textbf{How about scaling the inference time by majority voting?} Majority voting is an intuitive plug-and-play approach commonly used for scaling inference time by leveraging the collective reasoning process across different computational runs. 
To investigate the superimposed effect of majority voting and the aforementioned schemes, we conducted experiments with the Qwen2.5-72B model on the MedQA dataset. The results, illustrated in Figure~\ref{fig:mv_line_2}, compare the performance of four paradigms under majority voting (4, 8, 16, and 32 rounds).  Inference time was measured by the average number of output tokens per problem. Although Qwen2.5-72B-Vanilla shows a steady increase in performance with majority voting, the improvement is minimal, with accuracy rising only from 74.31\% to 74.63\%. Interestingly, when majority voting is combined with CoT reasoning (Qwen2.5-72B-Vanilla CoT), more improvements are observed. However, the accuracy reaches a peak of 80.44\% at higher token counts before slightly declining to 79.81\%. \textbf{Marginal gains of applying majority voting} are also observed in the proposed journey learning schemes (i.e., \texttt{LongStep} and \texttt{LongMonolog}), but more obvious than that in previous methods. Qwen2.5-72B-\texttt{LongStep} benefits a 1.26\% increment from majority voting while \texttt{LongMonolog} benefits 1.50\%. These results indicate that while majority voting can help refine predictions by aggregating outputs from multiple runs, it does not significantly enhance performance in tasks when intermediate steps are unthoughtful to achieve consistency in voting. Journey learning, which includes nuanced thoughts for reasoning, is a more hopeful way to enhance performance via majority voting.

\begin{figure}[h]
\centering
\scalebox{1}{
\includegraphics[width=\linewidth]{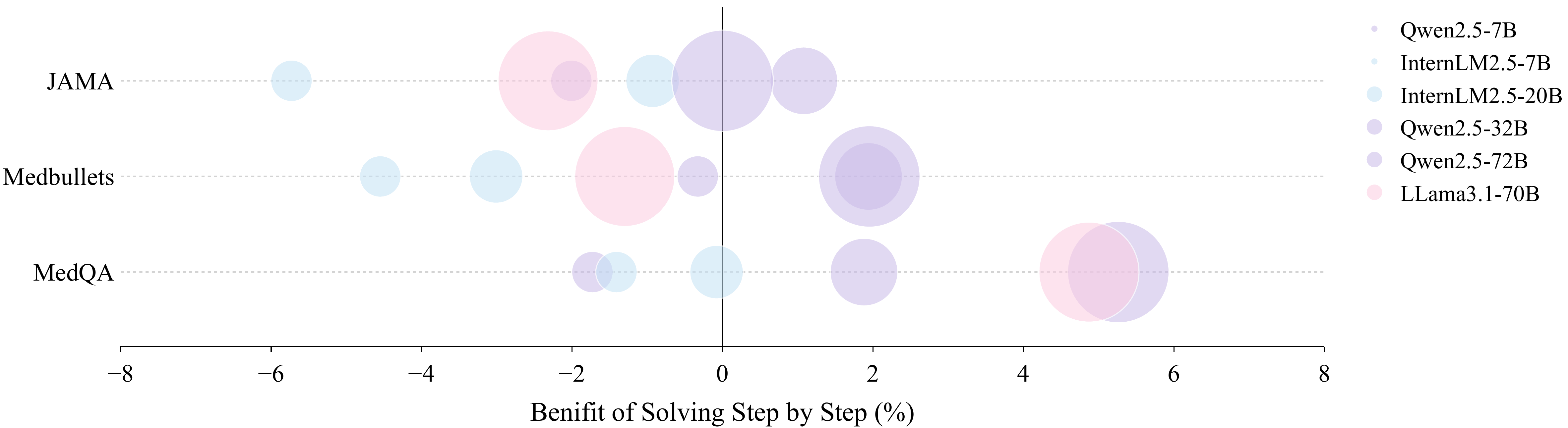}
}
\caption{The benefits of prompting open-source models to solve problems step by step are illustrated. The positive axis indicates that breaking down the problem into smaller steps can enhance model performance, while the negative axis suggests that doing so may lead to diminished returns. Each bubble represents a different model, with bubble size corresponding to the model's parameter size.}
\label{fig:cot_effect}
\end{figure}

As we stated at the beginning of this section, we only select models benefiting from vanilla CoT for further exploration. There is a question: \textbf{Does inference-time scaling always help?}  From this starting point, we first lay out the benefits of models using CoT prompting to solve medical problems of various difficulties. As vividly depicted in Figure~\ref{fig:cot_effect}, the models possessing a huge number of parameters are those achieving positive gains from CoT prompting. 
For models with smaller parameter sizes, such as 7B or 20B, increased inference time can unfortunately lead to performance degradation and, at times, failure to adhere to the instructed output format. On datasets of higher difficulty like JAMA, which contain challenging real-world clinical cases and require extensive domain knowledge for analysis, the performance deficits are particularly pronounced. Another noteworthy observation is that models with fewer parameters, such as Qwen2.5-32B, gain less from inference-time scaling than models with larger capacities. Specifically, Qwen2.5-32B and Qwen-72B respectively achieve increments of +1.66\% / +3.28\% (vanilla CoT), +2.93\% / +5.12\% (\texttt{CoT}), +5.88\% / +9.69\% (\texttt{LongStep}) and +6.03\% / +11.36\% (\texttt{LongMonolog}). Based on these findings, the underlying philosophy we hypothesize is: \textbf{The functioning of long thought during inference time requires sufficient capability,} otherwise a futile effort is expected. This is particularly significant in medicine, where solving clinical problems depends on the capacity to understand and generate complex and nuanced text and extensive knowledge, including aspects of diseases, pharmacology, and treatment protocols.

\begin{figure}[tb]
\centering
\scalebox{1}{
\includegraphics[width=\linewidth]{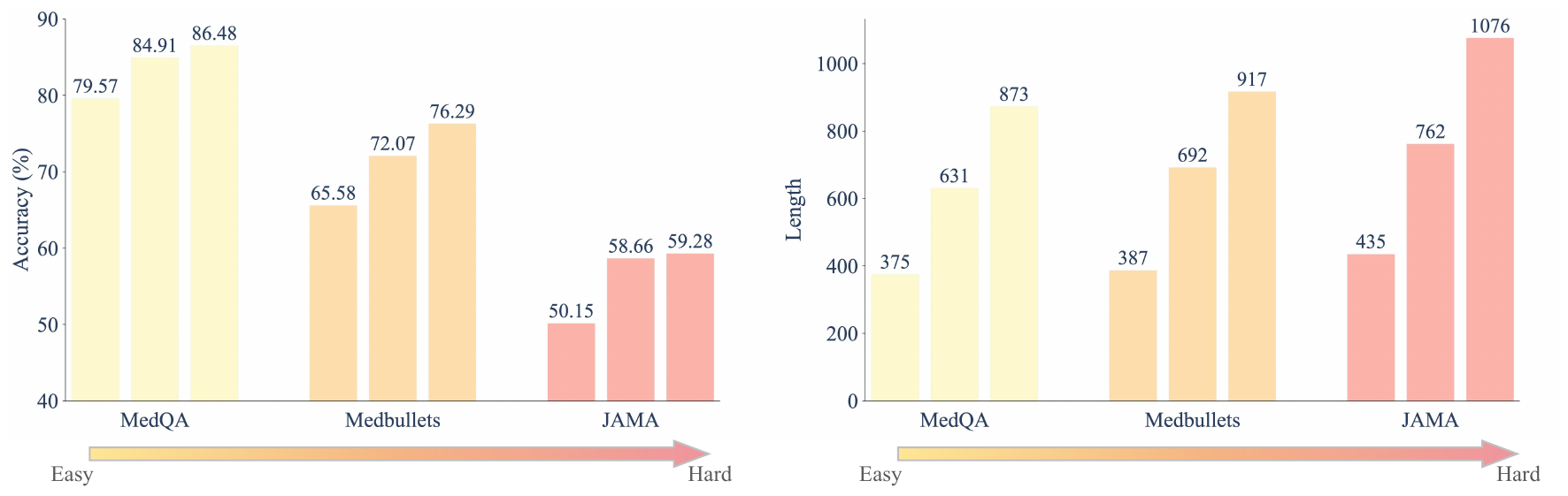}
}
\caption{Comparison of accuracy and average length of output tokens of Qwen2.5-72B across three datasets using distinct strategies(from left to right: Vanilla CoT, \texttt{LongStep} and \texttt{LongMonolog})}
\label{fig:acc_len}
\end{figure}

\subsection{Harder Tasks, Longer Thoughts, More Inference Time}
The right part of Figure~\ref{fig:acc_len} reveals an intriguing observation: harder tasks appear to require longer output tokens to benefit from inference-time computing.
To contextualize the level of difficulty, we hypothesized that answering questions in JAMA is more challenging than in Medbullets and MedQA, as JAMA presents more comprehensive real-world scenarios, and even proprietary models do not work well on JAMA. Additionally, we posited that Medbullets is more difficult than MedQA, as MedQA partially includes \textit{Step 1} questions from the USMLE. This hypothesis is partially validated by the overall performance of various settings presented on the left of Figure~\ref{fig:acc_len}, where MedQA achieves the highest accuracy, followed by Medbullets and JAMA. 

Examining the length of output tokens, Qwen2.5-72B utilizes an average of 1,076 tokens to answer questions in JAMA through detailed monolog, compared to 917 tokens in Medbullets and 873 tokens in MedQA. A similar increase in output length with task difficulty is evident across other reasoning paradigms and base models, as depicted in Table~\ref{tab:main}.
This philosophy is consistent with our initial observations outlined in Section~\ref{sec:exp_process_begin}, higher-difficulty questions necessitate more reasoning steps, thereby requiring more complex thinking and longer output during inference.

Another interesting observation is that models with fewer parameters, such as Qwen2.5-32B, tend to generate longer outputs when thinking aloud, particularly for easier questions. However, the opposite trend is observed in the other two reasoning paradigms, vanilla CoT prompting and long monolog SFT. Weaker models produce shorter responses in the former and outputs of similar length in the latter. After reviewing some examples, this discrepancy may stem from unnecessary verbosity when generating long monolog, as illustrated in \Cref{fig:case_problem_32,fig:case_step_32,fig:case_monolog_32}.~\footnote{\href{https://jamanetwork.com/journals/jamadermatology/fullarticle/2797659}{https://jamanetwork.com/journals/jamadermatology/fullarticle/2797659}} A weaker model may overlook or misinterpret key points during a long monolog, becoming stuck in confusion or arriving at incorrect conclusions. This finding, from the other side, supports the previous hypothesis that the functioning of inference-time scaling should be based on adequate knowledge.

\subsection{Generalizability and Future Directions}

When we take a closer look at the curated data in \texttt{LongStep} (in Figure~\ref{fig:case_step}) and \texttt{LongMonolog} dataset (in Figure~\ref{fig:case_monolog}), a surprising finding is that the data are not bounded by the options provided. The a priori options as input are internalized as heuristics for developing the output thought during inference time, which more resembles a complete diagnosis with differential candidates (and exclusion of them), rather than discussing the options sequentially. To verify whether models trained on journey learning data can be effective in the differential diagnosis context, we conducted a preliminary study. We removed the multiple-choice options and allowed the model to respond freely. To ensure fairness, we selected cases published in the 2024 JAMA Clinical Challenges while the training data were collected before October 2023. It is important to note that our model was exclusively trained on journey learning data, where the synthesis process included the provision of multiple-choice options. \Cref{fig:ddx,fig:ddx_qwen,fig:ddx_monolog} ~\footnote{\href{https://jamanetwork.com/journals/jamaoncology/fullarticle/2825256}{https://jamanetwork.com/journals/jamaoncology/fullarticle/2825256}} presents a representative case demonstrating the application of long-form reasoning for diagnosis. With long-form reasoning, the model tends to analyze a broader range of potential diseases while integrating various contextual information and its knowledge to approach more and more concise conclusions. These results offer promising insights for future research directions.

\section{Conclusion}
Building upon our initial exploration of inference-time scaling in the medical domain, our findings indicate that this approach offers promising enhancements for tackling complex reasoning tasks. This study demonstrated that \textit{inference-time scaling} significantly improves model performance across benchmarks like MedQA, Medbullets, and JAMA Clinical Challenges, with accuracy gains of 6–11\% achieved using only 500 training samples. Key insights at this stage primarily concern the effectiveness of scaling inference time. While majority voting offers a straightforward method to enhance inference-time computation, its impact remains limited compared to long reasoning paradigms. The necessity of extended reasoning for harder tasks underscores the scalability of inference-time scaling with task complexity. Furthermore, the shift from multiple-choice formats to free-form responses revealed the potential for nuanced medical journey learning, fostering deeper clinical reasoning capabilities.

Through continued exploration and iterative improvements, we aim to enhance the interpretability and effectiveness of inference-time scaling for addressing real-world medical challenges. By focusing on collaborative research and open resource sharing, we aim to strengthen the connection between computational advancements and real-world medical applications, ultimately improving diagnostic accuracy, patient outcomes, and healthcare efficiency. Our long-term goal is to develop intelligent systems that complement clinical expertise and address the growing complexity of modern medicine.

\begin{figure}[H]
\centering
\scalebox{1}{
\input{imgs/ddx}
}
\caption{Problem of Differential Diagnosis from JAMA Clinical Challenges}
\label{fig:ddx}
\end{figure}

\begin{figure}[H]
\centering
\scalebox{1}{
\input{imgs/ddx_qwen}
}
\caption{Free-form Response: Incorrect Output of Qwen2.5-72B to Differential Diagnosis}
\label{fig:ddx_qwen}
\end{figure}

\begin{figure}[H]
\centering
\scalebox{1}{
\input{imgs/ddx_monolog}
}
\caption{Free-form Respons: Correct Output of Qwen2.5-72B-\texttt{LongMonolog} to Differential Diagnosis}
\label{fig:ddx_monolog}
\end{figure}

\bibliographystyle{acl_natbib}
\bibliography{ref}
\newpage
\appendix
\section{Appendix}
\begin{figure}[h]
\centering
\scalebox{1}{
\includegraphics[width=\linewidth]{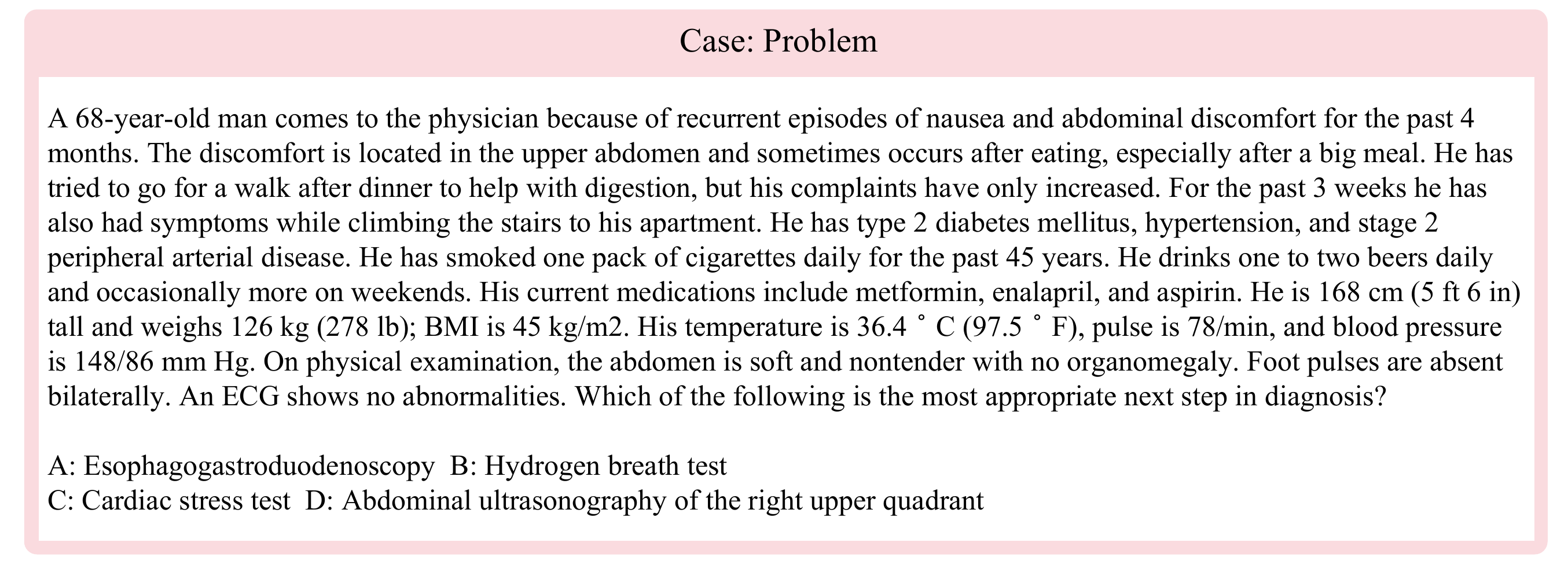}
}
\caption{Case of problems for synthesizing data.}
\label{fig:case_problem}
\end{figure}

\begin{figure}[h]
\centering
\scalebox{1}{
\includegraphics[width=\linewidth]{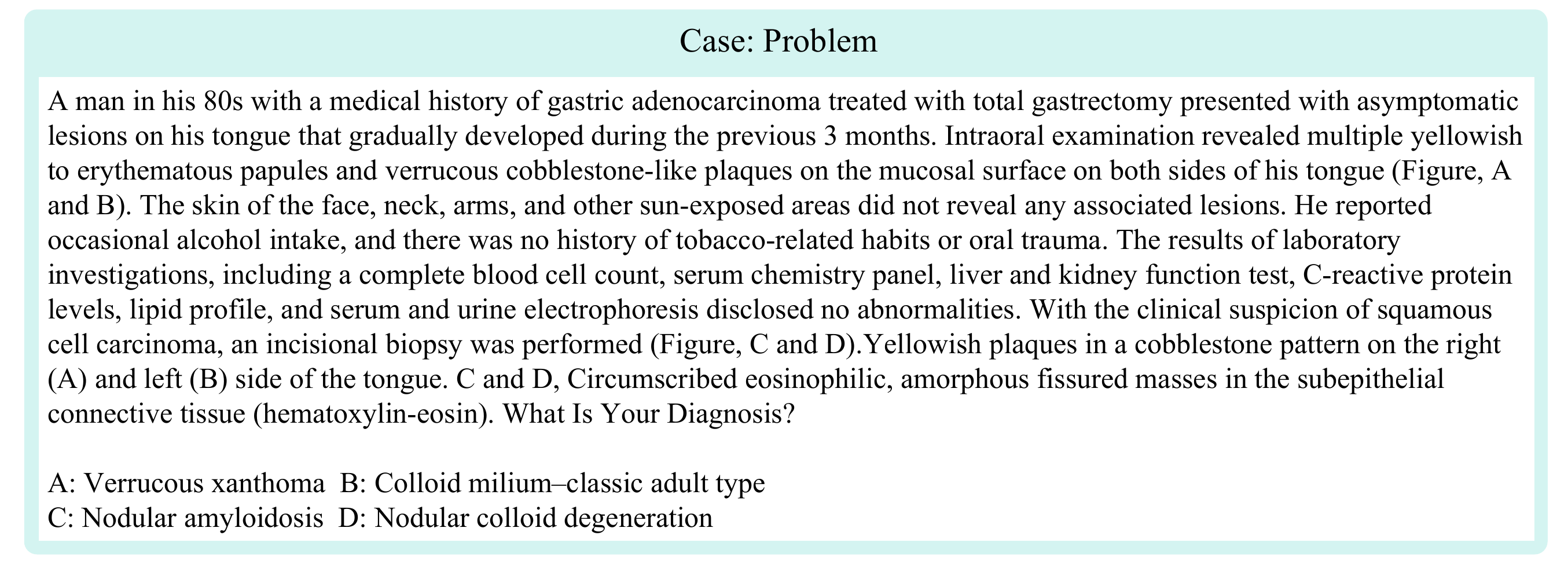}
}
\caption{One case of JAMA problems.}
\label{fig:case_problem_32}
\end{figure}

\begin{figure}[h]
\centering
\scalebox{1}{
\includegraphics[width=\linewidth]{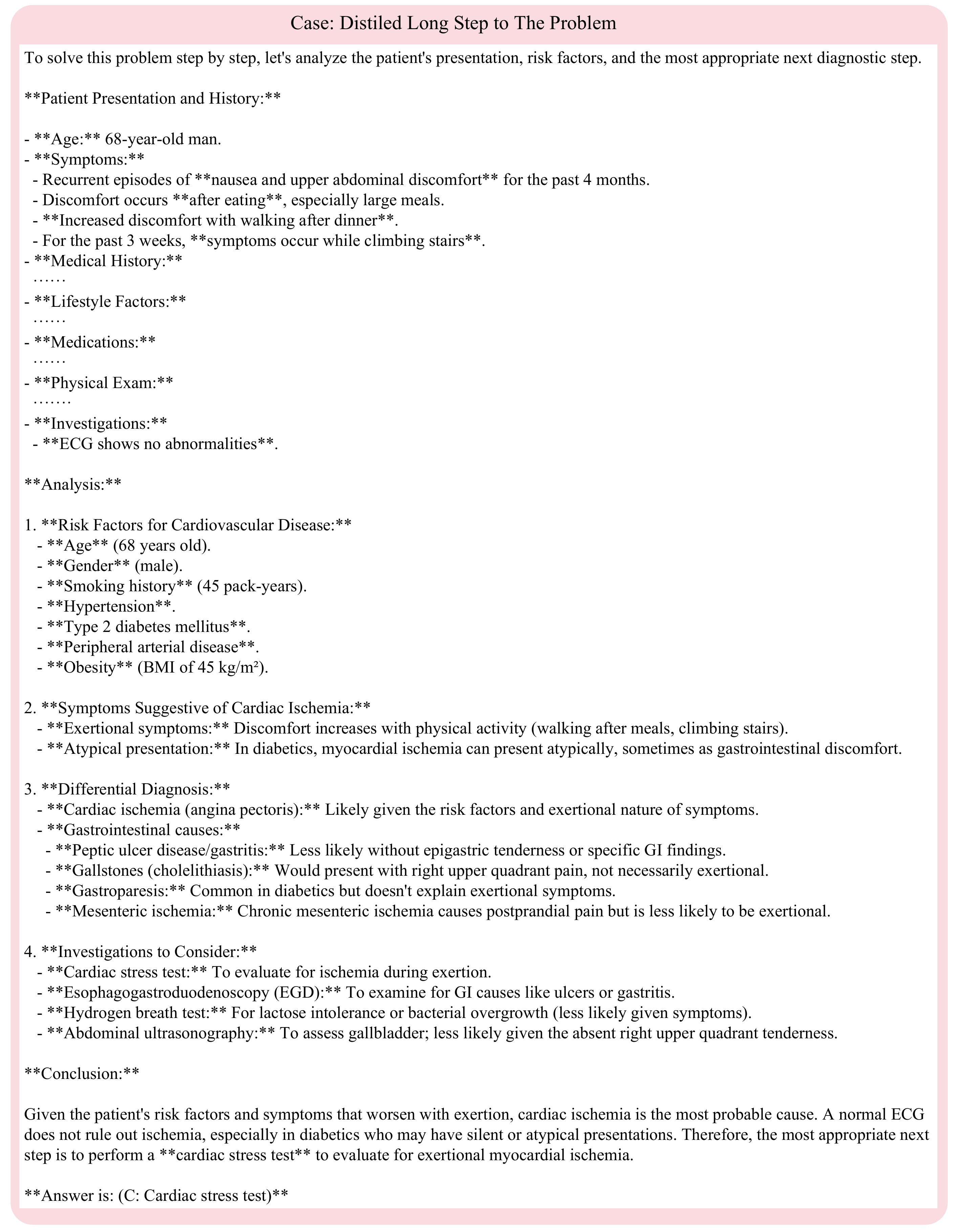}
}
\caption{Case of our distilled long step data for the problem.}
\label{fig:case_step}
\end{figure}

\begin{figure}[h]
\centering
\scalebox{1}{
\includegraphics[width=\linewidth]{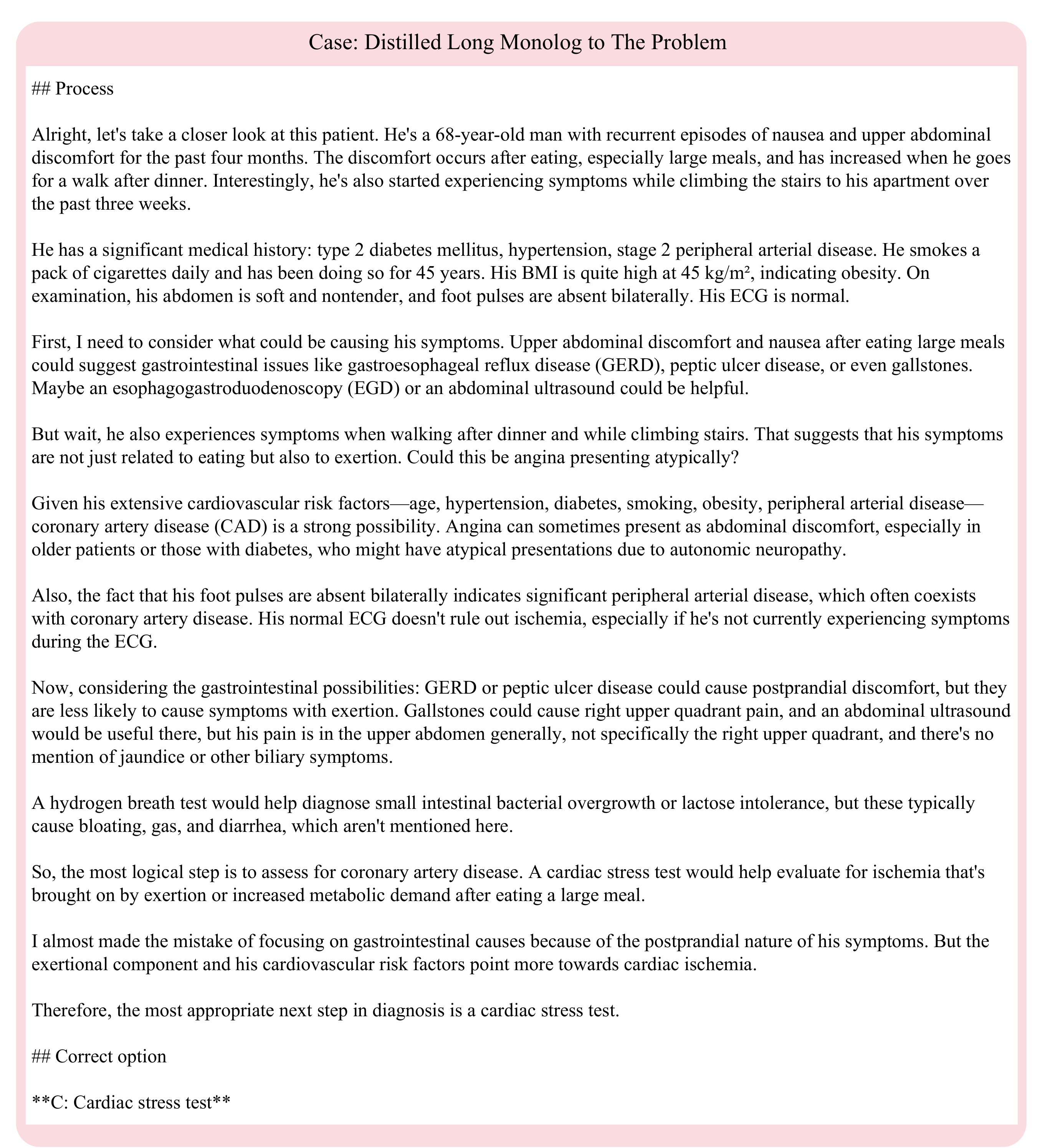}
}
\caption{One case of our distilled long Monolog data for the problem.}
\label{fig:case_monolog}
\end{figure}

\begin{figure}[h]
\centering
\scalebox{1}{
\includegraphics[width=\linewidth]{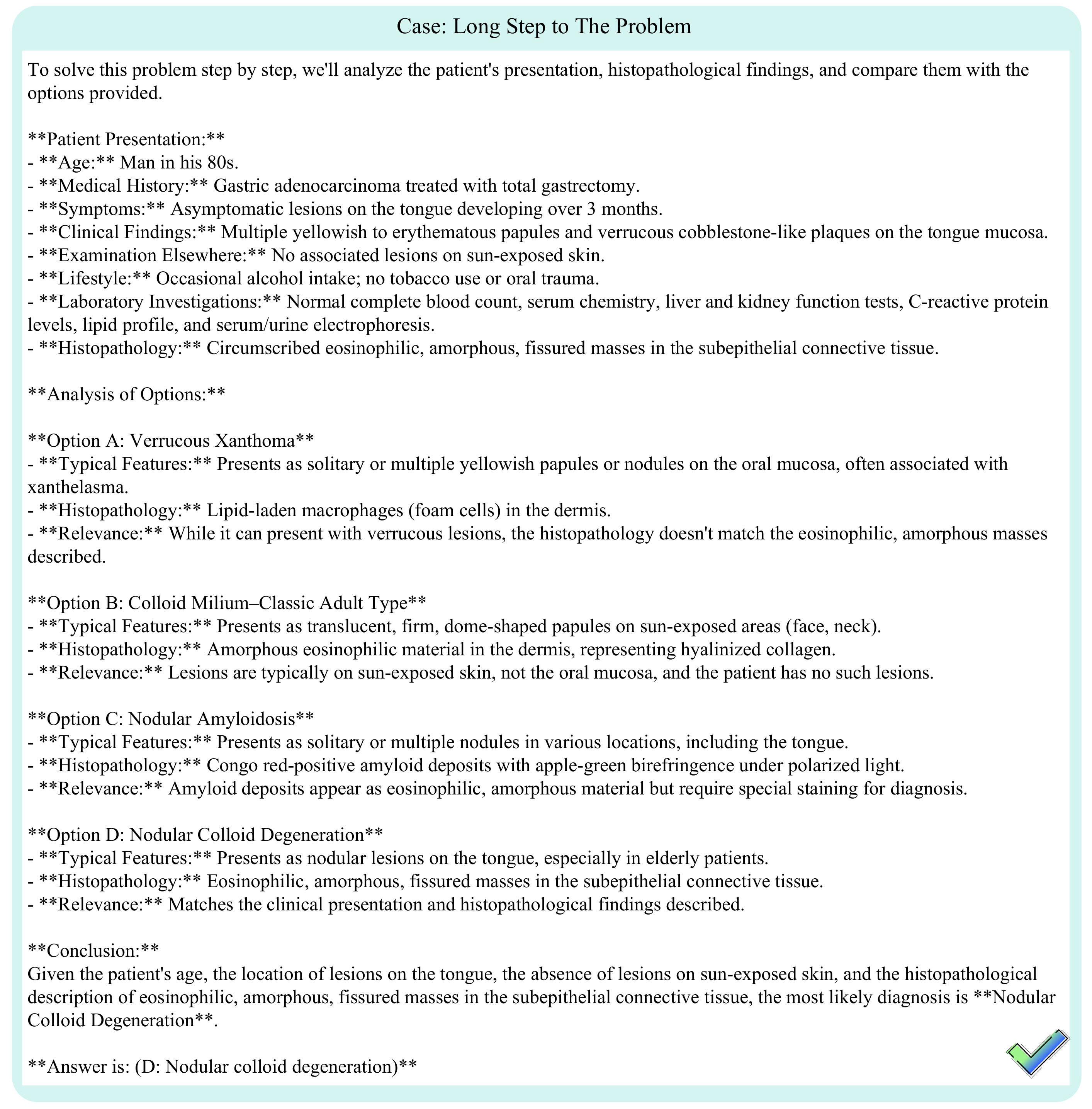}
}
\caption{Generated Thought by the Qwen2.5-32B-\texttt{LongStep} Model.}
\label{fig:case_step_32}
\end{figure}

\begin{figure}[h]
\centering
\scalebox{1}{
\includegraphics[width=\linewidth]{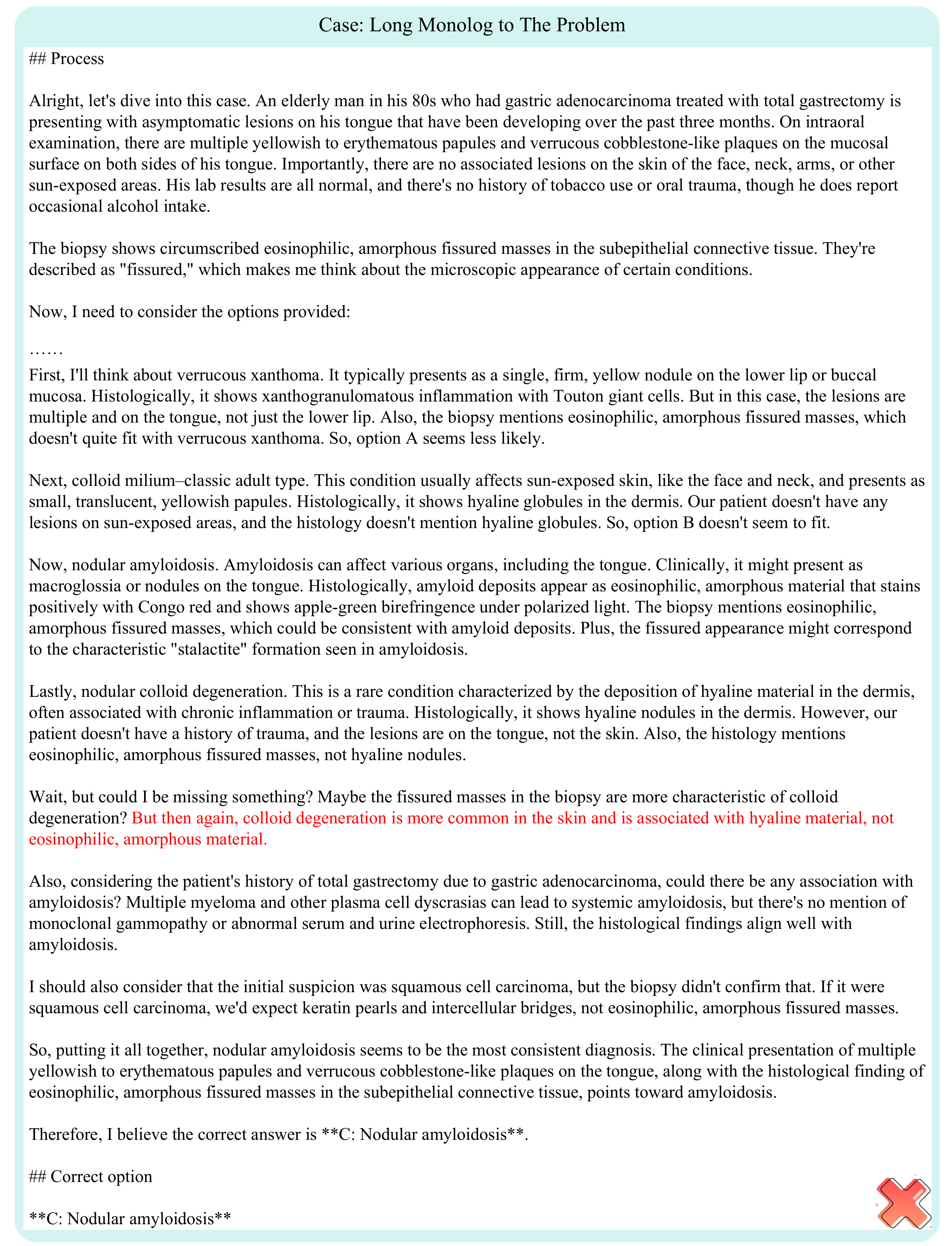}
}
\caption{Failure case generated by the Qwen2.5-32B-\texttt{LongStep} Model.}
\label{fig:case_monolog_32}
\end{figure}

\end{document}